\documentclass[letterpaper, 10 pt, conference, twoside]{IEEEtran}

\usepackage{times}
\usepackage{epsfig}
\usepackage{graphicx}
\usepackage{amsmath}
\usepackage{amssymb}
\usepackage{booktabs}
\usepackage{multirow}
\usepackage{xcolor}
\usepackage{balance}
\usepackage{makecell}

\usepackage{bm}
\usepackage{subcaption}
\usepackage{mathtools}

\newcommand{\hide}[1]{}

\newcommand{\bdmath}{\begin{dmath}}
\newcommand{\edmath}{\end{dmath}}
\newcommand{\beq}{\begin{equation}}
\newcommand{\eeq}{\end{equation}}
\newcommand{\bdm}{\begin{displaymath}}
\newcommand{\edm}{\end{displaymath}}
\newcommand{\bea}{\begin{eqnarray}}
\newcommand{\eea}{\end{eqnarray}}
\newcommand{\beal}{\beq \begin{array}{ll}}
\newcommand{\eeal}{\end{array} \eeq}
\newcommand{\beas}{\begin{eqnarray*}}
\newcommand{\eeas}{\end{eqnarray*}}
\newcommand{\ba}{\begin{array}}
\newcommand{\ea}{\end{array}}
\newcommand{\bit}{\begin{itemize}}
\newcommand{\eit}{\end{itemize}}
\newcommand{\ben}{\begin{enumerate}}
\newcommand{\een}{\end{enumerate}}

\newcommand{\SO}{\mathrm{SO}}
\newcommand{\so}{\mathfrak{so}}
\newcommand{\SE}{\mathrm{SE}}

\newcommand{\Real}{\mathbb{R}}

\newcommand{\Identity}{\mathbf{I}}

\newcommand{\residual}{\mathbf{r}}

\newcommand{\Zero}{\mathbf{0}}

\newcommand{\tran}{\mathbf{p}}

\newcommand{\vel}{\mathbf{v}}

\newcommand{\gravity}{\mathbf{g}}

\newcommand{\expmap}{\mathrm{Exp}}
\newcommand{\logmap}{\mathrm{Log}}

\newcommand{\mat}[1]{\mathbf{#1}}
\newcommand{\matsym}[1]{\bm{#1}}
\renewcommand{\vec}[1]{\mathbf{#1}}
\newcommand{\vecsym}[1]{\bm{#1}}

\newcommand{\rot}{\mat{R}}

\usepackage[breaklinks=true,bookmarks=false]{hyperref}

\IEEEoverridecommandlockouts

\title{\huge \vspace{0.5cm} Visual-Inertial Mapping with Non-Linear Factor Recovery}

\author{
Vladyslav Usenko$^1$,
Nikolaus Demmel$^1$,
David Schubert$^1$, 
J\"org St\"uckler$^2$ and 
Daniel Cremers$^1$ %

\thanks{$^{1}$ Vladyslav Usenko, Nikolaus Demmel, David Schubert and Daniel Cremers are with the Technical University of Munich, Germany {\tt\footnotesize \{usenko, demmeln, schubdav, cremers\}@in.tum.de}}%
\thanks{$^{2} $ J\"org St\"uckler is with MPI for Intelligent Systems T\"ubingen, Germany {\tt\footnotesize joerg.stueckler@tuebingen.mpg.de}}
}

\begin{document}


\maketitle

\begin{abstract}

    Cameras and inertial measurement units are complementary sensors for ego-motion estimation and environment mapping. 
    Their combination makes visual-inertial odometry (VIO) systems more accurate and robust. 
    For globally consistent mapping, however, combining visual and inertial information is not straightforward. 
    To estimate the motion and geometry with a set of images large baselines are required. 
    Because of that, most systems operate on keyframes that have large time intervals between each other. 
    Inertial data on the other hand quickly degrades with the duration of the intervals and after several seconds of integration, it typically contains only little useful information.
    
    In this paper, we propose to extract relevant information for visual-inertial mapping from visual-inertial odometry using non-linear factor recovery. We reconstruct a set of non-linear factors that make an optimal approximation of the information on the trajectory accumulated by VIO. To obtain a globally consistent map we combine these factors with loop-closing constraints using bundle adjustment. The VIO factors make the roll and pitch angles of the global map observable, and improve the robustness and the accuracy of the mapping.
    In experiments on a public benchmark, we demonstrate superior performance of our method over the state-of-the-art approaches.
\end{abstract}

\section{Introduction}

Visual-inertial odometry  (VIO) is a popular approach for tracking the motion of a camera in application domains such as robotics or augmented reality. 
By combining visual and IMU measurements, one can exploit the complementary strengths of both sensors and thereby increase accuracy and robustness.
Commonly, the optimization of camera trajectory and map is performed locally on a small window of recent camera frames and IMU measurements.
This approach, however, is inevitably prone to drift in the estimates.

Globally consistent optimization for visual-inertial mapping is less explored in the computer vision community.
While in principle the optimization could be formulated as bundle adjustment with additional IMU measurements, this approach would quickly become computationally infeasible due to the high number of frames which would lead to a large number of optimization parameters in a naive formulation.
To keep the computational burden in bounds, bundle adjustment subsamples the high-frame rate images of the camera to a smaller set of keyframes.
The common choice in VIO is to preintegrate IMU measurements between consecutive frames.
If we select keyframes temporally far apart to make the optimization efficient, the preintegrated IMU measurements provide only little information to constrain the trajectory due to the accumulated sensor noise. 
The small frame rate also affects the quality of the estimated velocities and biases from visual and inertial cues which are required for pose prediction using preintegrated IMU measurements.

\begin{figure}[t]
    \centering

    \includegraphics[width=\linewidth]{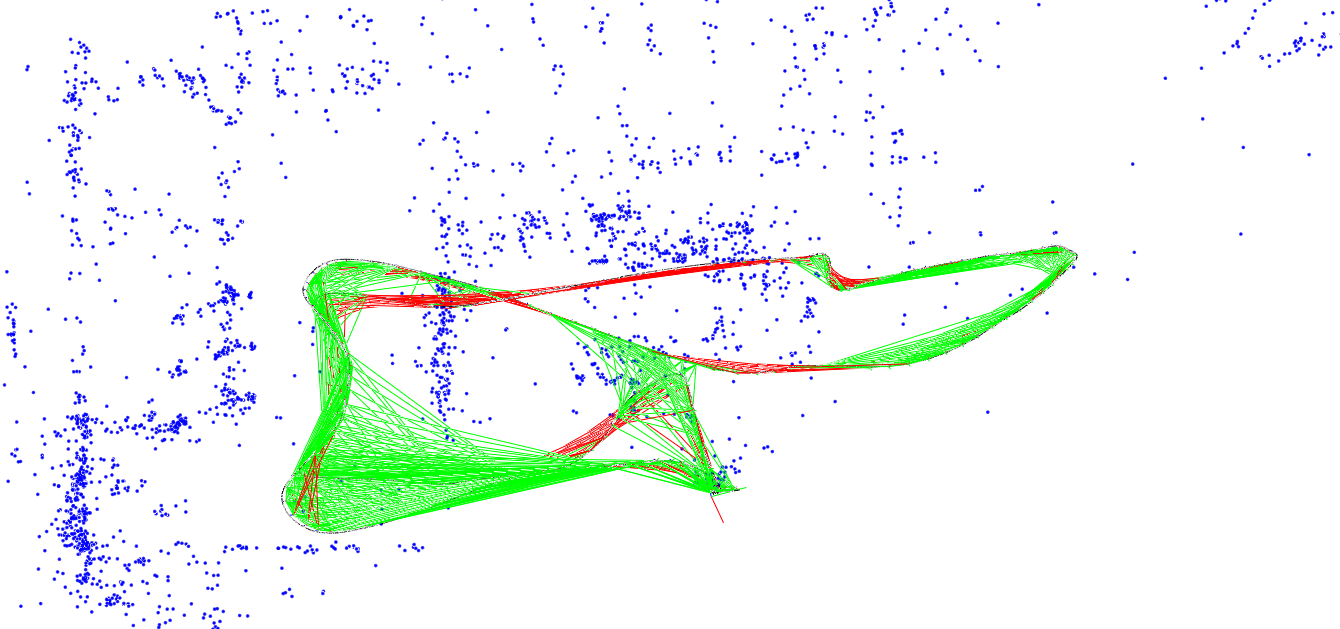}

    \caption{Orthographic top-down projection of the map (MH\_05 sequence of the EuRoC dataset~\cite{Burri16}) rendered using the estimated gravity direction. To obtain a gravity-aligned globally consistent map, non-linear factors are recovered from the marginalization prior of the VIO and combined with keypoint-based bundle adjustment. Green lines visualize keyframe connections resulting from bundle adjustment factors and red lines connections from the recovered relative pose factors. Additionally each keyframe has a recovered factor that penalizes deviation from the gravity direction observed in VIO. }
    \label{fig:teaser}
    \vspace{-5mm}
\end{figure}

We propose a novel approach that formulates visual-inertial mapping as bundle adjustment on a high-frame-rate set of visual and inertial measurements.
Instead of directly optimizing the camera trajectory for all frames, we propose a hierarchical approach which first recovers a local VIO estimate at the frame rate of the camera.
Once keyframes are removed and marginalized from the current local VIO optimization window, we extract non-linear factors~\cite{mazuran2016nonlinear} that approximate the accumulated visual-inertial information about the camera motion between keyframes.
The keyframes and non-linear factors are subsequently used on the global bundle-adjustment layer.

For the VIO layer, our method uses image features designed for fast and accurate tracking, while for the mapping layer we employ distinctive but lighting and viewpoint invariant keypoints that are suitable for loop closing. 
With this, our approach can leverage information from the IMU and short-term visual tracking at high frame rates together with keypoint matching and loop-closing at low frame rates for globally consistent mapping (Fig. \ref{fig:teaser}).
The factors also help to keep the map gravity-aligned, bridge between frames that do not have enough visual information. Our approach also makes the optimization problem smaller, since we do not have to estimate velocities and biases. 

In summary, our contributions are:
\begin{itemize}
    \item We propose a novel two-layered visual-inertial mapping approach that integrates keypoint-based bundle-adjustment with inertial and short-term visual tracking through non-linear factor recovery.
    \item As the first layer of our mapping approach we propose a VIO system which outperforms the state-of-the-art methods in terms of trajectory accuracy on the majority of the evaluated sequences.
    This is achieved by carefully combining appropriate components (patch tracking, landmark representation, first-estimate Jacobians, marginalization scheme) as detailed in Sec. \ref{sec:vio}.
    \item Unlike other state-of-the-art systems that use preintegrated IMU measurements also for mapping, we subsume high-frame rate visual-inertial information in non-linear factors extracted from the marginalization prior of the VIO layer. This results not only in a smaller optimization problem but also in better pose estimates in the resulting gravity aligned map.
\end{itemize}

We encourage the reader to watch the demonstration video and
inspect the open-source implementation of the system, which is available at:

\begin{center}
\color{purple}
\urlstyle{tt}
\textbf{\footnotesize \url{https://vision.in.tum.de/research/vslam/basalt}}
\end{center}

\section{Related Work}

{\bf Visual-inertial odometry: } 
Early methods for visual-inertial odometry are primarily filter-based~\cite{jones11ijrr, Mourikis2007a}.
In tightly integrated filters, the prediction step typically propagates the current camera state estimate using the IMU measurements. 
The state is recursively corrected based on the camera images.
A significant drawback of filters is that the linearization point for the non-linear measurement and state transition models cannot be changed, once a measurement is integrated.
Fixed-lag smoothers (a.k.a. optimization-based approaches) such as~\cite{leutenegger2014keyframe, usenko2016icra} relinearize at the current states in a local optimization window of recent frames.
The visual-inertial state estimation is formulated as a full bundle adjustment~(BA) over keyframes and IMU measurements.
The problem is reduced to a computationally manageable size by marginalization of old frames up to the recent set in the optimization window.
The continuous relinearization, windowed optimization and maintenance of the marginalization prior increase the accuracy of the methods.
The above methods need to discard keypoints and observations that are observed in marginalized keyframes in order to maintain the sparse structure of the marginalization prior.
Hsiung et al.~\cite{hsiung2018} apply non-linear factor recovery to achieve a sparse marginalization prior without discarding information about observed keypoints. 
This way, the approach can further refine the keypoints and achieve higher accuracy, but in contrast to our work it is limited to local BA.

{\bf Visual-inertial mapping: } 
Only few works have tackled globally consistent mapping from visual and inertial measurements.
Kasyanov et al.~\cite{Kasyanov2017_VISLAM} add a pose-graph optimization layer with loop-closing on top of a keyframe-based visual-inertial odometry method~\cite{leutenegger2014keyframe}.
The pose graph is built from the keyframes of the VIO and their relative pose estimates.
In~\cite{mur-artal-inertial}, the authors add inertial measurements to a keyframe-based SLAM system through IMU preintegration. 
The IMU measurements are preintegrated into a set of pseudo-measurements between keyframes. 
They notice that the accuracy of preintegrated measurements degrades over time and restrict the time between keyframes to 0.5 seconds in local BA and 3 seconds in global BA. 
A further shortcoming of the method is its requirement of estimating the camera velocity and IMU biases at each keyframe which is less well constrained through visual measurements than in our approach due to the strong temporal subsampling into keyframes.
Schneider et al.~\cite{schneider2018maplab} follow a similar approach in which preintegrated IMU measurements are inserted into the optimization.
The approach in \cite{qin2019} proposes a combination of VIO and 4 degree-of-freedom~(DoF) pose optimization for visual-inertial mapping. 
They fix 2 DoF (roll and pitch) and optimize only for the others. 
We also constrain roll and pitch from visual-inertial measurements.
However, we extract non-linear factors in a probabilistic formulation which account for uncertainties in those values and are traded off with other information in the global probabilistic optimization.

\section{Preliminaries}
	
	In this paper, we write matrices as bold capital letters (e.g.\ $\mat{R}$) and vectors as bold lowercase letters (e.g.\ ${\vecsym{\xi}}$). 
	Rigid-body poses are represented as $ (\rot, \tran) \in  \SO(3) \times \Real^3 $ or as transformation matrices $\mat{T} \in \SE(3)$ when needed.
	Incrementing a rotation $\rot$ by an increment $\vecsym{\xi} \in \Real^3$ is defined as $\rot \oplus \vecsym{\xi} = \expmap(\vecsym{\xi}) \rot$. 
	The difference between two rotations $\rot_1$ and $\rot_2$ is calculated as  $\rot_1 \ominus \rot_2 = \logmap(\rot_1 \rot_2^{-1})$  such that $(\rot \oplus \vecsym{\xi}) \ominus \rot = \vecsym{\xi}$. 
	Here we use $\expmap\colon\Real^3\rightarrow\SO(3)$, which is a composition of the hat operator ($\Real^3\rightarrow\so(3)$) and the matrix exponential ($\so(3)\rightarrow\SO(3)$) and maps rotation vectors to their corresponding rotation matrices, and its inverse $\logmap\colon\SO(3)\rightarrow\Real^3$. For all other variables, such as translation, velocity and biases, we define $\oplus$
    and $\ominus$ as regular addition and subtraction.
	
	In the following we will use a state $\vec{s}$ that is defined as a tuple of several rotation and vector variables, and a function $\vec{r}(\vec{s})$ that depends on it and can also produce rotations and vectors as the result.
	An increment $\vecsym{\xi} \in \Real^{n}$ is a stacked vector with all the increments of the variables in $\vec{s}$. Then, the Jacobian of the function with respect to the increment is defined as
	\begin{align}\label{eq:jac}
		\mat{J}_{\vec{r}(\vec{s})} =  \lim_{\vecsym{\xi}\rightarrow\vec{0}}\frac{ \vec{r}(\vec{s} \oplus \vecsym{\xi}) \ominus \vec{r}(\vec{s})} {\vecsym{\xi}}\,.
	\end{align}
	Here, $\vec{s} \oplus \vecsym{\xi}$ denotes that each component in $\vec{s}$ is incremented with the corresponding segment in $\vecsym{\xi}$ using the appropriate definition of the $\oplus$ operator, and similarly for $\ominus$. 
	The limit is done component-wise, such that the Jacobian is a matrix. For Euclidean quantities, this definition is just a normal derivative, with an extension for rotations, both as function value and as function argument.
	For more details and possible alternative formulations we refer the reader to \cite{barfoot2017state, bloesch16calculus, eade2014lie}.
	
	In non-linear least squares problems, we minimize functions of the form
    \begin{align}
        \label{eq:gnerr}
        E(\vec{s}) = \frac{1}{2} \vec{r}(\vec{s})^{\top} \mat{W} \vec{r}(\vec{s})\,,
    \end{align}
    which is a squared norm of the sum of residuals with block-diagonal weight matrix $\mat{W}$. In this case, $\vec{r}(\vec{s})$ is purely vector-valued. Near the current state $\vec{s}$ we can use a linear approximation of the residual, which leads to
	\begin{align}
	    E(\vec{s}\oplus \vecsym{\xi}) = E(\vec{s}) +  \vecsym{\xi}^{\top} \mat{J}_{\vec{r}(\vec{s})}^{\top} \mat{W} \vec{r}(\vec{s})  + \frac{1}{2} \vecsym{\xi}^{\top} \mat{J}_{\vec{r}(\vec{s})}^{\top} \mat{W}  \mat{J}_{\vec{r}(\vec{s})} \vecsym{\xi}\,.
	\end{align}
	The optimum of this approximated energy can be attained using the Gauss-Newton increment
	\begin{align}
	\label{eq:gninc}
	\vecsym{\xi}^{*} =  -(\mat{J}_{\vec{r}(\vec{s})}^{\top} \mat{W}  \mat{J}_{\vec{r}(\vec{s})})^{-1} \mat{J}_{\vec{r}(\vec{s})}^{\top} \mat{W} \vec{r}(\vec{s})\,.
	\end{align}
	With this, we can iteratively update the state $\vec{s}_{i+1} = \vec{s}_{i} \oplus \vecsym{\xi}^{*}$ until convergence.

\section{Visual-Inertial Odometry}
\label{sec:vio}

We formulate the incremental motion tracking of the camera-IMU setup over time as fixed-lag smoothing. First, we use patch-based optical flow to track a sparse set of points in the 2D image plane between consecutive frames. This information is then used in a bundle-adjustment framework which for every frame minimizes an error that consists of point reprojection and IMU propagation terms. To maintain a fixed parameter size of the optimization problem we marginalize out old states. In the remainder of this section we will discuss these stages in more detail.

\subsection{KLT Tracking}
As a first step of our algorithm we detect a sparse set of keypoints in the frame using the FAST \cite{fast10} corner detector. To track the motion of these points over a series of consecutive frames we use sparse optical flow based on KLT \cite{lukas_kanade}. To achieve fast, accurate and robust tracking we combine the inverse-compositional approach as described in \cite{matthews01} with a patch dissimilarity norm that is invariant to intensity scaling. Several authors suggested zero-normalized cross-correlation (ZNCC) for illumination-invariant optical flow \cite{molnar10, Steinbruecker-et-al-vmv09}, but we use locally-scaled sum of squared differences (LSSD) defined in \cite{roma2002comparative} which is computationally less expensive than alternatives.

\begin{figure}[t]
    \centering
    \includegraphics[width=\linewidth]{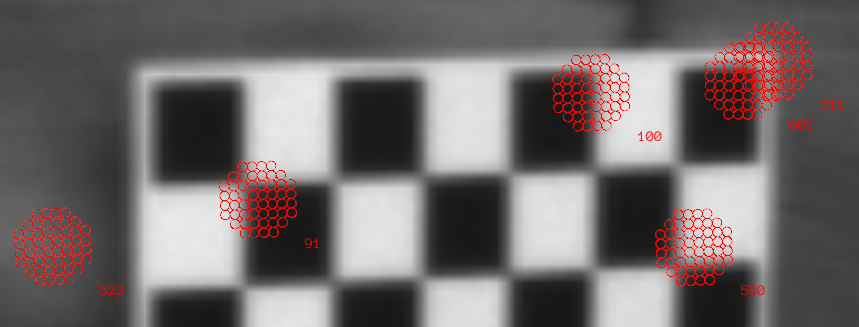}
    \includegraphics[width=\linewidth]{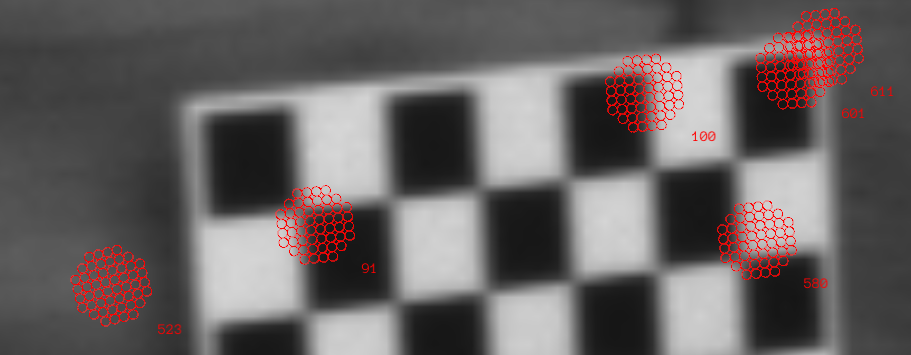}
    \caption{Example of KLT tracks estimated by our system. Despite changes in exposure time the proposed method is able to estimate the warp in SE(2) between the patches in the images. }
    \vspace{-5mm}
    \label{fig:flow}
\end{figure}

We formulate the patch tracking problem as estimating the transform $\mat{T} \in \SE(2)$ between two corresponding patches in two consecutive frames that minimizes the differences between the patches according to the selected norm. Essentially, we minimize a sum of squared residuals, where every residual is defined as
\begin{align}
\label{eq:flow}
    r_{i}(\vecsym{\xi}) &=   \frac{I_{t+1}(\mat{T} \mathbf{x}_i)}{\overline{I_{t+1}}} - \frac{I_{t}(\mathbf{x}_i)}{ \overline{I_{t}}} \quad \forall \mathbf{x}_i \in \Omega.
\end{align}
 Here, $I_t(\vec{x})$ is the intensity of image $t$ at pixel location $\vec{x}$. The set of image coordinates that defines the patch is denoted $\Omega$ and the mean intensity of the patch in image $t$ is $\overline{I_{t}}$. A visualization of the patch and tracking results is shown in Fig.~\ref{fig:flow}.

To achieve robustness to large displacements in the image we use a pyramidal approach, where the patch is first tracked on the coarsest level and then on increasingly finer levels. For outlier filtering, instead of an absolute threshold on the error, we track the patches from the current frame to the target frame and back to check consistency. Points that do not return to the initial location with the second tracking are considered as outliers and discarded.

\subsection{Visual-Inertial Bundle Adjustment}

To estimate the motion of the camera we combine error terms based on tracked feature locations from KLT tracking with IMU error terms based on preintegrated IMU measurements \cite{forster15}. 

We use the following coordinate frames throughout the paper: $\text{W}$ is the world frame, $\text{I}$ is the IMU frame and $\text{C}_i$ is the frame of camera $i$, where $i$ is the index of the camera in a stereo setup. We estimate transformations $\mat{T}_\text{WI} \in \SE(3)$ from IMU to world coordinate frame. The transformations $\mat{T}_{\text{IC}_i}$ from camera frame $i$ to IMU frame and the projection functions $\pi_i$ are assumed to be static and known from calibration. For the formulation of reprojection errors we denote the transformations from camera $i$ to world by $\mat{T}_{\text{WC}_i}$. Those do not constitute additional optimization variables and are calculated using $\mat{T}_\text{WI}$ and $\mat{T}_{\text{IC}_i}$ in practice.

At different points in time, we optimize a state
\begin{align}
    \vec{s} = \{\vec{s}_\text{k}, \vec{s}_\text{f}, \vec{s}_\text{l}\}\,,
\end{align}
where $\vec{s}_\text{k}$ contains IMU poses for $n$ older keyframes, $\vec{s}_\text{f}$ contains IMU poses, velocities and biases of the $m$ most recent frames, which possibly are also keyframes if they host landmarks, and $\vec{s}_\text{l}$ contains landmarks. A graphical representation of the problem is shown in Fig. \ref{fig:factor_graph} (a). Landmarks are stored relative to the keyframe where they were observed for the first time \cite{mei2011rslam} and defined by a unit-length direction vector in the coordinate frame of the camera and an inverse distance to the landmark \cite{civera2008inverse}. In the proposed system only keyframes host landmarks, which distinguishes them from regular frames.

\subsubsection{Representation of Unit Vectors in 3D}
\label{sec:unit_3d}

In order to avoid the necessity of additional constraints for the optimization and to keep the number of optimiziation variables small, we parametrize the bearing vector in 3D space using a minimal representation, which is two-dimensional. In \cite{bloesch2017_rovio} the authors provide an extensive review of possible parametrizations and suggest a new parametrization based on $\SO(3)$ rotations that yields simple derivatives with respect to 2D increments.

\begin{figure}[t]
    \centering
    \includegraphics[width=0.5\linewidth]{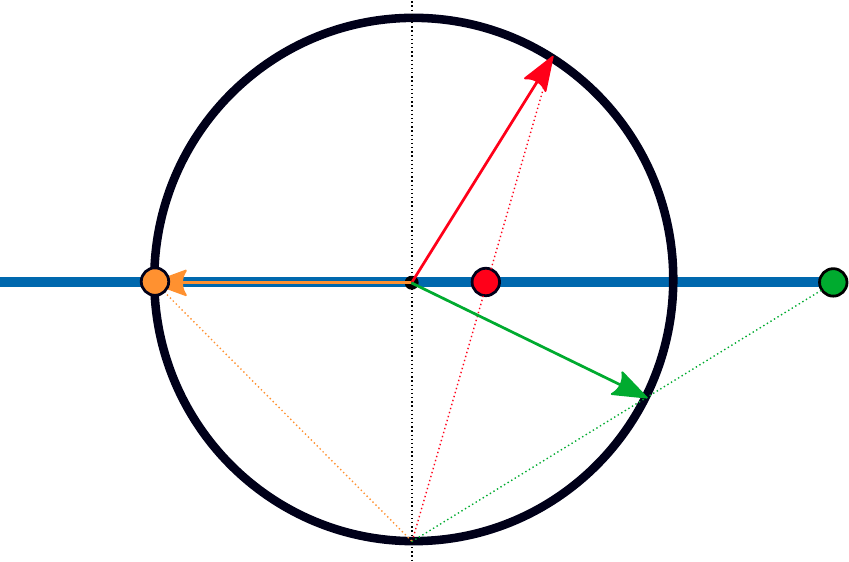}
    \caption{Geometric interpretation of stereographic projection used to represent unit vectors. 
    The two parameters define a point in the $XY$-plane of the coordinate system shown in blue. To obtain the corresponding 3D unit vector we cast a ray from $(0 ~~ 0 ~ -1)^{\top}$ and find an intersection with the unit sphere shown in black. Three example points are visualized in red, green and yellow, with dashed lines representing the rays intersecting with the sphere and arrows showing the resulting unit vectors.}
    \vspace{-5mm}
    \label{fig:unit_param}
\end{figure}

In this work we use a parametrization based on stereographic projection that given 2D coordinates $(u,v)^{\top}$  generates a unit-length bearing vector
\begin{align}
   \begin{pmatrix}
    x \\
    y \\
    z 
\end{pmatrix} &=  \begin{pmatrix}
    \eta u \\
    \eta v \\
    \eta - 1 
\end{pmatrix}\,,\quad
   \eta = \frac{2}{1 + u^2 + v^2}\,.
\end{align}
This parametrization is efficient as it only uses simple operations such as multiplication and division (compared to trigonometric operations needed in \cite{civera2008inverse}) and is defined for all $u$ and $v$. 
A geometric interpretation is shown in Fig.~\ref{fig:unit_param}. 
The only direction vector that cannot be represented with finite $u,v$ is the negative $Z$-direction $\begin{pmatrix}0 & 0 & -1\end{pmatrix}^{\top}$. However, this is not a drawback in practice, as cameras usually have a limited field of view and cannot see points behind them.

\begin{figure*}[t]
    \centering
    \includegraphics[width=0.3\linewidth]{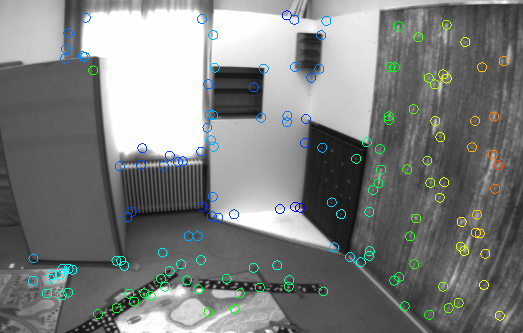}
    \includegraphics[width=0.3\linewidth]{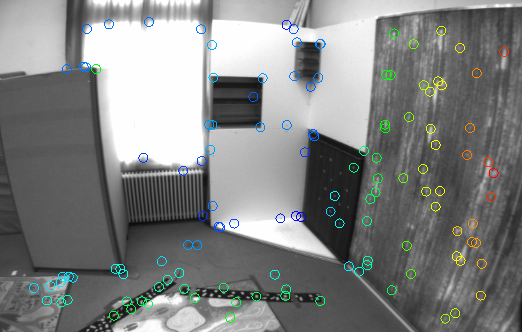}
    \includegraphics[width=0.38\linewidth]{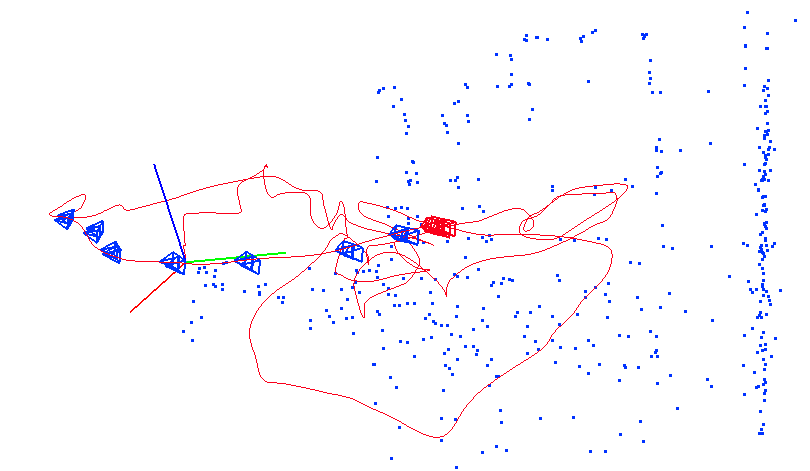}
      \caption{Visual-inertial odometry subsystem proposed in Section \ref{sec:vio}. Projections of the landmarks with color-coded inverse distance used for estimating the position of the current frame are shown on the left. The results of local visual-inertial bundle adjustment are shown on the right. Keyframe poses with the associated landmarks are visualized in blue, current states and the estimated trajectory are visualized in red. Information about the keyframe poses in the local window is approximated using a set of non-linear factors as described in Section \ref{sec:vi_mapping} and reused for global mapping. }
      \vspace{-5mm}
    \label{fig:vio}
\end{figure*}

\subsubsection{Reprojection Error}

The first cue we can use for motion estimation is the reprojection error. When point $i$ that is hosted in frame $h(i)$ is detected in target frame $t$ at image coordinates $\vec{z}_{it}$, the residual is defined as
\begin{align}
   \label{eq:reproj}
   \vec{r}_{it} &= \vec{z}_{it} - \pi_{c(t)}( \mat{T}_{t}^{-1} \mat{T}_{h(i)} \vec{q}_i(u,v,d))\,, \\
   \mathbf{q}_i(u,v,d) &= \begin{pmatrix}
    x(u,v) &
    y(u,v) &
    z(u,v) &
    d
\end{pmatrix}^\top\,,
\end{align}
where $c(t)$ is the index of the camera used to take frame~$t$. The pose $\mat{T}_{t}$ denotes $\mat{T}_{\text{WC}_{c(t)}}$ at the time when frame~$t$ has been taken, and similarly for $\mat{T}_{h(i)}$. The first three entries of the homogeneous point coordinates $\vec{q}_i(u,v,d)$ are computed from the minimal representation $(u,v)$ as described in Sec.~\ref{sec:unit_3d}, with an additional fourth entry $d$, the inverse distance. Since the projection function is independent of scale we do not have to normalize $\vec{q}_i$, which makes this formulation numerically stable even when $d$ is close or equal to zero.

\subsubsection{IMU Error}

The second cue for motion estimation is the IMU data. To deal with the high frequency of IMU measurements we preintegrate several consecutive IMU measurements into a pseudo-measurement. When adding an IMU factor between frame $i$ and frame $j$, we compute pseudo-measurement $\Delta\vec{s}=(\Delta\rot, \Delta\vel, \Delta\tran)$ similar to~\cite{forster15}. For this, we compute bias-corrected accelerations $\vec{a}_t=\vec{a}_t^\text{raw}-\bar{\vec{b}}_i^\text{a}$ and rotational velocities $\vecsym{\omega}_t=\vecsym{\omega}_t^\text{raw}-\bar{\vec{b}}_i^\text{g}$ using the raw accelerometer $\vec{a}_t^\text{raw}$ and gyroscope $\vecsym{\omega}_t^\text{raw}$ measurements. We fix the corresponding biases $\bar{\vec{b}}_i^\text{a}$ and $\bar{\vec{b}}_i^\text{g}$ for the entire preintegration time and use linear approximation to account for changes in these variables.

For the timestamp $t_i$ of frame $i$, we assign the initial state delta $\Delta\vec{s}_{t_i}=(\Identity,\Zero,\Zero)$. Then, for each IMU timestamp $t$ satisfying $t_i<t\leq t_j$ the following updates are calculated. 
%
\begin{align}
    \Delta\rot_{t+1} &= \Delta\mat{R}_{t} \expmap(\vecsym{\omega}_{t+1}\Delta t)\,,\\
    \Delta\vel_{t+1} &= \Delta\vel_{t} +  \Delta\mat{R}_{t} \vec{a}_{t+1}\Delta t\,,\\
    \Delta\tran_{t+1} &= \Delta\tran_{t} + \Delta\vel_{t} \Delta t \,.
    \label{eq:propend}
\end{align}
This defines $\Delta\vec{s}_{t+1}$ as a function of $\Delta\vec{s}_t$, $\vec{a}_{t+1}$,  and $\vecsym{\omega}_{t+1}$,
\begin{align}
    \Delta\vec{s}_{t+1} = f(\Delta\vec{s}_t,\vec{a}_{t+1}, \vecsym{\omega}_{t+1})\,,
\end{align}
with corresponding Jacobian $\mat{J}_f=[\mat{J}_f^{\text{s}},\mat{J}_f^\text{a},\mat{J}_f^{\text{g}}]$.
Furthermore, all previous iterations of $f$ up to $t+1$ define $\Delta\vec{s}_{t+1}$ as a function of the biases,
\begin{align}
    \Delta\vec{s}_{t+1} = g_{t+1}(\vec{b}_i^\text{a}, \vec{b}_i^\text{g})\,.
\end{align}
Starting with zero-initialization, the corresponding Jacobian $\mat{J}_{g_{t+1}}=[\mat{J}_{g_{t+1}}^\text{a}, \mat{J}_{g_{t+1}}^\text{g}]$ can be computed recursively using $\mat{J}_f$,
\begin{align}
    \mat{J}_{g_{t+1}}^\text{a}&=\mat{J}_f^{\text{s}}\mat{J}_{g_{t}}^\text{a} - \mat{J}_f^\text{a}\,,\\
    \mat{J}_{g_{t+1}}^\text{g}&=\mat{J}_f^{\text{s}}\mat{J}_{g_{t}}^\text{g} - \mat{J}_f^\text{g}\,,
\end{align}
which results from the chain rule. Eventually, the Jacobians of $g_{t_j}$ are denoted $\mat{J}^\text{g}$ and $\mat{J}^\text{a}$. 
Small changes in biases can be represented as increments to the linearization point $\vec{b}^\text{a}_i=\bar{\vec{b}}^\text{a}_i+\vecsym{\epsilon}^\text{a}$ and $\vec{b}^\text{g}_i=\bar{\vec{b}}^\text{g}_i+\vecsym{\epsilon}^\text{g}$. Then, $\Delta\vec{s}$ is approximated as
\begin{align}
    \Delta\tilde{\vec{s}}(\vec{b}^\text{a}_i, \vec{b}^\text{g}_i) = \Delta\vec{s}(\bar{\vec{b}}^\text{a}_i, \bar{\vec{b}}^\text{g}_i) \oplus (\mat{J}^\text{a}\vecsym{\epsilon}^\text{a} + \mat{J}^\text{g}\vecsym{\epsilon}^\text{g})\,,
\end{align}
with components $\Delta\tilde{\vec{s}}=(\Delta\tilde{\rot}, \Delta\tilde{\vel}, \Delta\tilde{\tran})$.
The residuals are then calculated as
\begin{align}
    \residual_{\Delta\rot} &= \logmap\left(\Delta\tilde{\rot}\rot_j^\top \rot_i\right)  \,,\\
    \residual_{\Delta\vel} &= \rot_i^\top(\vel_j - \vel_i - \gravity\Delta t) - \Delta\tilde{\vel}\,,\\
    \residual_{\Delta\tran} &= \rot_i^\top (\tran_j - \tran_i - \frac{1}{2}\gravity\Delta t^2) - \Delta\tilde{\tran}\,,
\end{align}
where $\gravity$ is the gravity vector and $\rot$ and $\tran$ denote the rotation and translation components of $\mat{T}_\text{WI}$, respectively. These residuals have to be weighted with an appropriate covariance matrix, which can be also calculated recursively. Starting from $\matsym{\Sigma}_{t_i}=\mat{0}$, updates are calculated as
\begin{align}
    \matsym{\Sigma}_{t+1}
    = \mat{J}_f^\text{s} \matsym{\Sigma}_{t} {\mat{J}_f^\text{s}}^\top
    + \mat{J}_f^\text{a}\matsym{\Sigma}^\text{a}{\mat{J}_f^\text{a}}^\top
    + \mat{J}_f^\text{g}\matsym{\Sigma}^\text{g}{\mat{J}_f^\text{g}}^\top\,,
\end{align}
with diagonal matrices $\matsym{\Sigma}^\text{a}$ and $\matsym{\Sigma}^\text{g}$ that contain the hardware-specific IMU noise parameters for accelerometer and gyroscope. 
For more detailed information about the underlying physical model of the IMU and preintegration theory we refer the reader to \cite{forster15}.

\subsubsection{Optimization and Partial Marginalization}

For each new frame we minimize a non-linear energy that consists of reprojection terms, IMU terms and a marginalization prior $E_\text{m}$
\begin{align}
    E &= \sum_{\mathclap{\substack{i\in\mathcal{P}\\t\in\mathrm{obs}(i)}}} \vec{r}_{it}^\top \matsym{\Sigma}^{-1}_{it} \vec{r}_{it} + \sum_{\mathclap{\substack{(i,j)\in\mathcal{C}}}} \vec{r}_{ij}^\top \matsym{\Sigma}^{-1}_{ij} \vec{r}_{ij} + E_\text{m}.
\end{align}
The reprojection errors are summed over the set of points $\mathcal{P}$ and for each point $i$ over the set $\mathrm{obs}(i)$ of frames where the point is observed, including its host frame. The set $\mathcal{C}$ contains pairs of frames which are connected by IMU factors.

The energy $E$ is optimized using the Gauss-Newton algorithm. To constrain the problem size we fix the number of keyframe poses and consecutive states that we optimize at every iteration. When a new frame is added, there are $n$ pose-only keyframes in $\vec{s}_\text{k}$ and the $m$ newest frames including the newly added one in $\vec{s}_\text{f}$. After optimizing, we perform a partial marginalization of the state to prevent the problem size from growing.

\begin{figure*}[t]
    \centering
    \subcaptionbox{}{\includegraphics[height=0.23\linewidth]{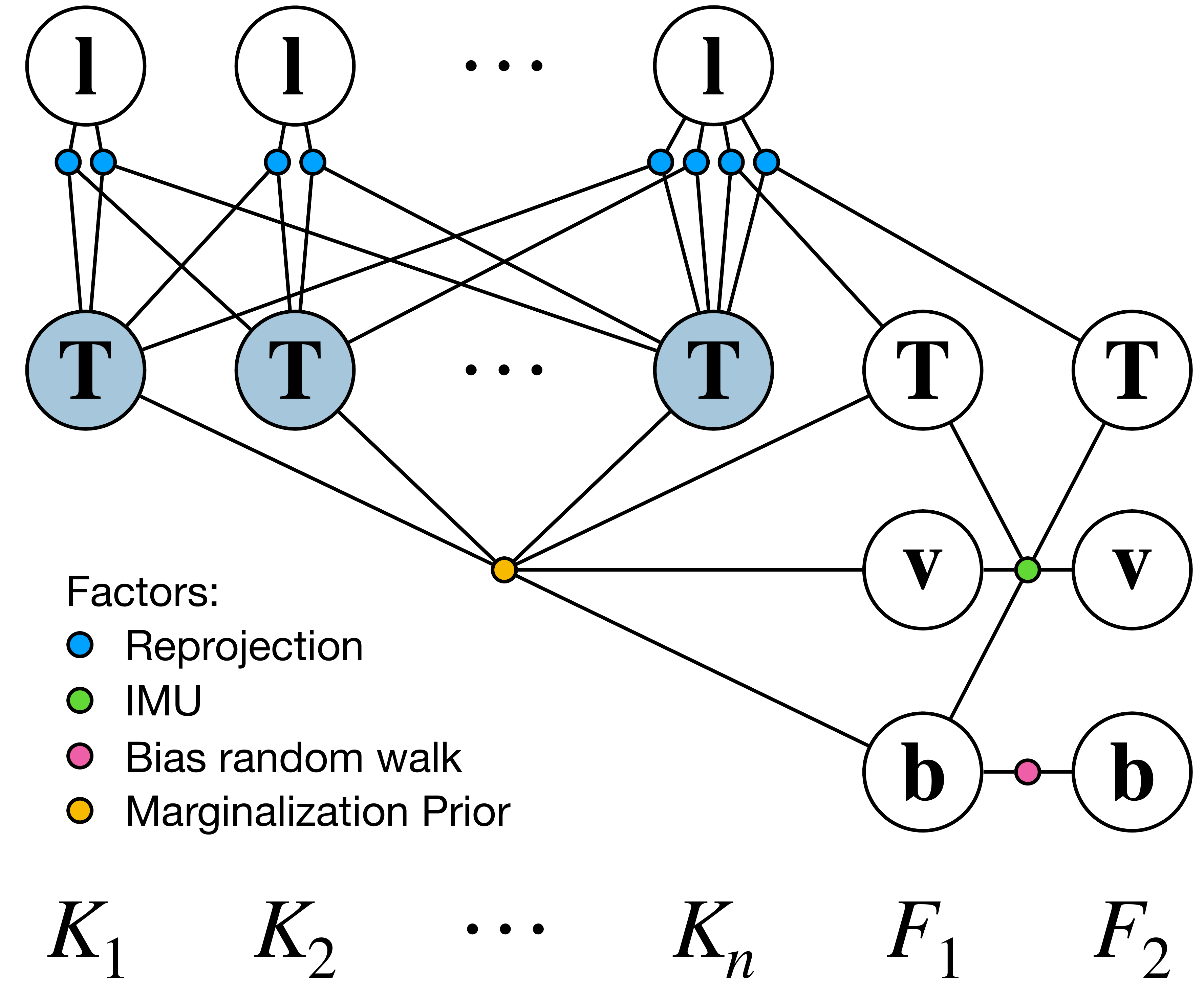}}\hfill
    \subcaptionbox{}{\includegraphics[height=0.23\linewidth]{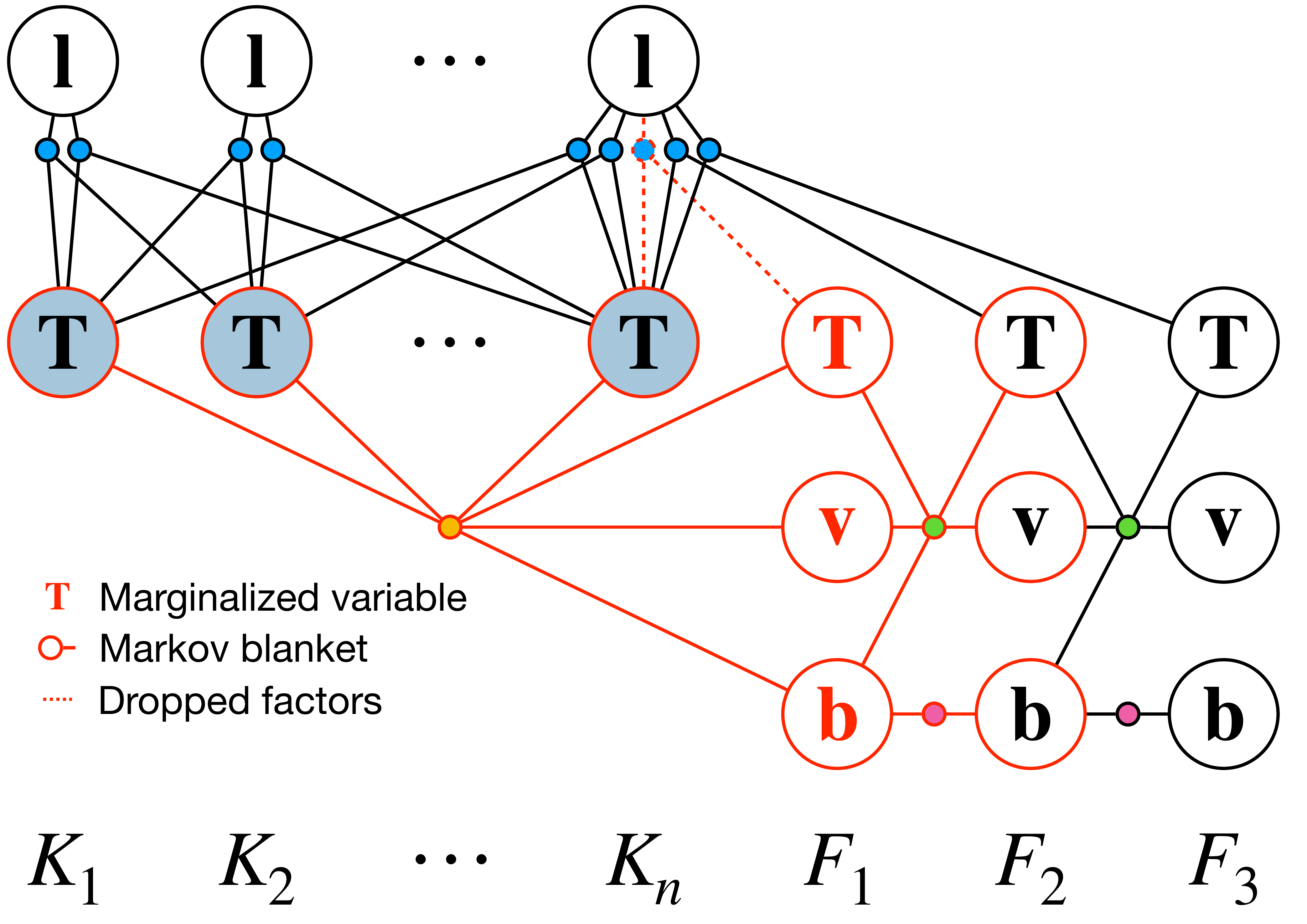}}\hfill
    \subcaptionbox{}{\includegraphics[height=0.23\linewidth]{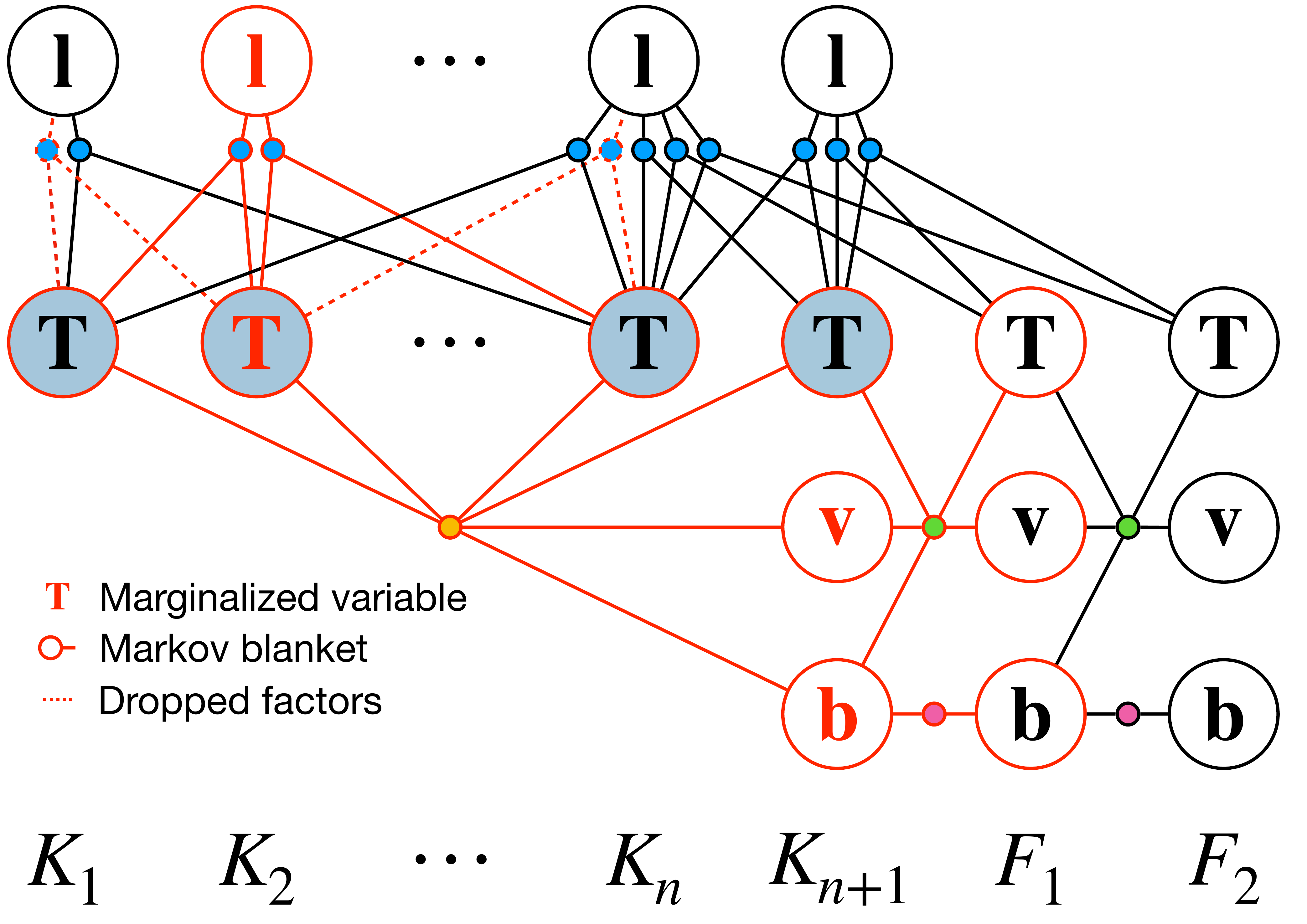}}
    \caption{Factor graphs. (a) After marginalizing a frame, the system consists of $n$ older keyframes $K_1\dots K_n$ and the $m-1$ most recent frames $F_1$ and $F_2$ (which could potentially also host landmarks and hence be keyframes). After a new frame has been added, the oldest velocity $\vec{v}$ and the oldest bias $\vec{b}$ are marginalized. If they do not belong to a keyframe (b), the whole frame including its pose $\vec{T}$ is marginalized. If they belong to a keyframe (c), another keyframe is selected for marginalization, including the landmarks hosted in it and its pose. In both cases, reprojection factors where the target frame is the marginalized frame are dropped. In the latter case, reprojection factors from the marginalized frame to $F_2$ are dropped to allow relinearization. Note that not all possible combinations of host and target frames for reprojection factors are shown.}
    \vspace{-3mm}
    \label{fig:factor_graph}
\end{figure*}

Two possible scenarios for marginalization are shown in Fig. \ref{fig:factor_graph}. In the first one we marginalize out the oldest non-keyframe. In this case we drop the landmark factors that have this frame as a target to maintain the sparsity of the problem. In the second case we have a new keyframe, so we marginalize out velocity and biases for this frame and one old keyframe with corresponding landmarks.

In both cases the marginalization is done on the linearized Markov blanket of the variables we want to remove, where the Markov blanket is a collection of incident states to those variables. The linearization $\mat{H}$ and $\vec{b}$ represent a distribution of the estimated state in the vector space of the increment~$\vecsym{\xi}$.
If we split the increment $\vecsym{\xi}=[\vecsym{\xi}_\alpha^\top, \vecsym{\xi}_\beta^\top]^\top$ into variables $\vecsym{\xi}_\alpha$ to stay in the system and variables $\vecsym{\xi}_\beta$ to be marginalized, 
we can compute the parameters of the new distribution using the Schur complement,
\begin{align}
    \mat{H}^\text{m}_{\alpha\alpha}&=\mat{H}_{\alpha\alpha} - \mat{H}_{\alpha\beta}\mat{H}_{\beta\beta}^{-1}\mat{H}_{\beta\alpha}\,,\\
    \vec{b}^\text{m}_{\alpha}&=\vec{b}_{\alpha} - \mat{H}_{\alpha\beta}\mat{H}_{\beta\beta}^{-1}\vec{b}_\beta\,,
\end{align}
where we have split the original $\mat{H}$ and $\vec{b}$ into
\begin{align}
    \mat{H}=
    \begin{bmatrix}
        \mat{H}_{\alpha\alpha} & \mat{H}_{\alpha\beta} \\
        \mat{H}_{\beta\alpha} & \mat{H}_{\beta\beta}
    \end{bmatrix}\,,\quad
    \vec{b}=
    \begin{bmatrix}
        \vec{b}_\alpha \\
        \vec{b}_\beta
    \end{bmatrix}\,.
\end{align}
$\mat{H}^\text{m}_{\alpha\alpha}$ and $\vec{b}^\text{m}_{\alpha}$ now define an energy term that only depends on $\vecsym{\xi}_\alpha$ and can be added to the total energy at the next iteration.
	
We use first-estimate Jacobians \cite{huang2009first} to maintain the nullspace properties of the linearized marginalization prior. As soon as a variable becomes a part of the marginalization prior, its linearization point is fixed, and the Jacobian used to calculate $\mat{H}$ and $\vec{b}$ is evaluated at this linearization point, while the residuals are calculated at the current state estimate. Residuals already in the marginalization term have to be linearly approximated, thus not $\vec{b}_\alpha^\text{m}$, but $\vec{b}_\alpha^\text{m}+\mat{H}^\text{m}_{\alpha\alpha}\vecsym{\delta}_\alpha$ is added to the Gauss-Newton optimization once $\vecsym{\xi}_\alpha$ deviates by $\vecsym{\delta}_\alpha$ from the state used to calculate the residuals in $\vec{b}_\alpha^\text{m}$.

\section{Visual-Inertial Mapping}
\label{sec:vi_mapping}

The fixed-lag smoothing method for visual-inertial odometry (Fig. \ref{fig:vio}) presented in the previous section accumulates drift in the estimate due to the fixed linearization points outside the optimization window. 
A typical approach to eliminate such drift is to detect loop closures and incorporate loop-closing constraints into the optimization.
We propose a two-layered approach which runs our visual-inertial odometry on the lower layer and bundle-adjustment on the visual-inertial mapping layer, where we additionally use non-linear factors that summarize the keyframe pose information from the odometry layer.
BA optimizes the camera poses of keyframes and positions of keypoints.
We implicitly detect loop closures using keypoint matching and achieve globally consistent mapping.

\begin{figure}[t]
    \centering
    \includegraphics[width=\linewidth]{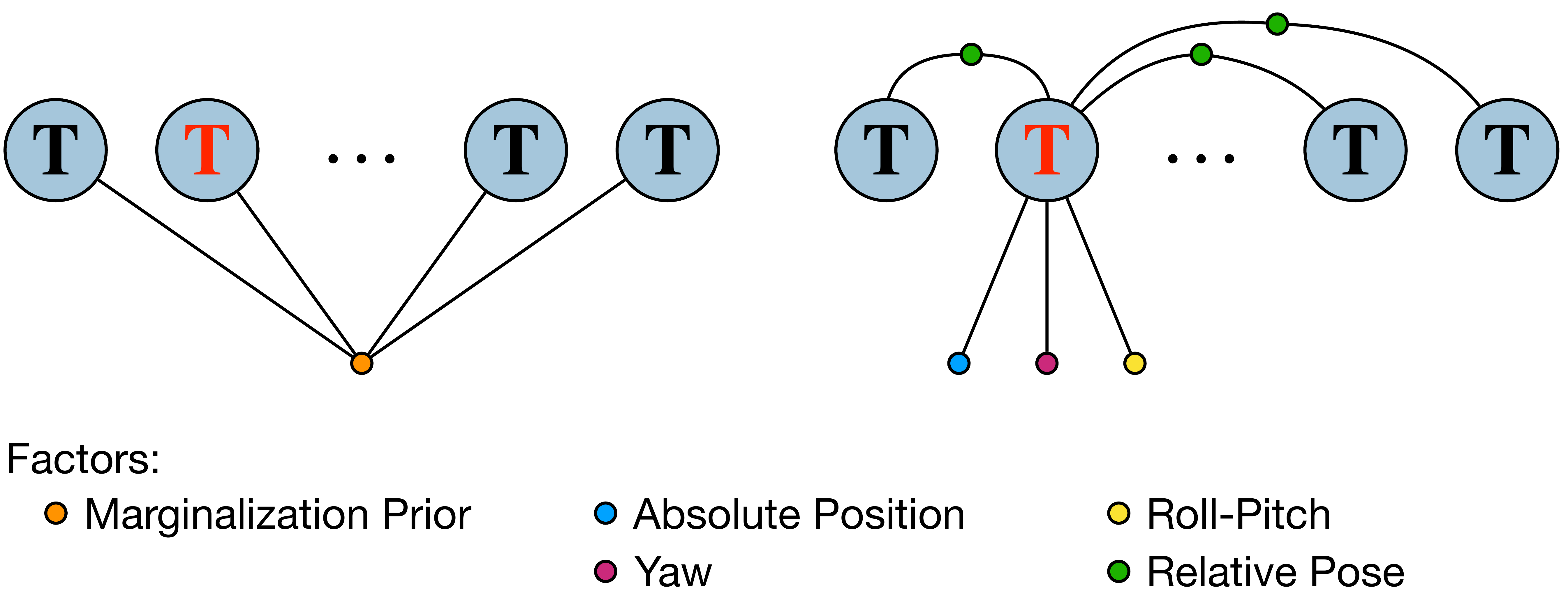}
    \caption{Visualization of non-linear factor recovery. Left: Densely connected factor from marginalization saved from the VIO before removing a keyframe pose. Right: Extracted non-linear factors that approximate the distribution stored in the original factor.}
    \vspace{-5mm}
    \label{fig:recovered_graph}
\end{figure}

\subsection{Global Map Optimization}

To get statistically independent observations we detect and match ORB~\cite{rublee11} features (distinct from VIO points) between the keyframes in the global map optimization. 
This allows us to use the reprojection error function as defined in Eq.~\eqref{eq:reproj}.
Combining this reprojection error with the error terms from the recovered non-linear factors yields the objective function:
\begin{align}
    E^\text{G}(\vec{s}) &= \sum_{\mathclap{\substack{i\in\mathcal{P}\\t\in\mathrm{obs}(i)}}} 
    \vec{r}_{it}^\top \matsym{\Sigma}^{-1}_{it} \vec{r}_{it} + E_{\text{nfr}}(\vec{s}),
\end{align}
where $E_{\text{nfr}}(\vec{s})$ collects the error terms by the recovered non-linear factors. 
These factors and their recovery are detailed in the following.
The state~$\vec{s}$ that we optimize on this global optimization layer includes the keyframe poses and the positions of the new landmarks (parametrized as in Sec.~\ref{sec:unit_3d}).

We interface the global map optimization with the VIO layer at the keyframe poses.
When a keyframe is marginalized out from the VIO we save the linearization of the Markov blanket (Fig. \ref{fig:factor_graph} (c)) and marginalize all other variables except of keyframe poses. From this marginalization prior, we recover a set of non-linear factors on the keyframe poses that approximate the distribution stored in it.

\subsection{Non-Linear Factor Recovery}
\label{sec:nonlinear_factor}

Non-linear factor recovery (NFR~\cite{mazuran2016nonlinear}) approximates a dense distribution stored in the linearized Markov blanket of the original factor graph with a different set of non-linear factors that yield a sparse factor graph topology.
While the initial aim of NFR is to keep the computational complexity of SLAM optimization bounded, we use it to transfer information accumulated during VIO to our globally consistent visual-inertial map optimization. 

By linearization of the residual function of a non-linear least squares problem Eq.~\eqref{eq:gnerr}, we obtain a multivariate Gaussian distribution $p(\vec{s}) \sim N(\vecsym{\mu}_\text{o}, \mat{H}_\text{o}^{-1})$ in which the mean $\vecsym{\mu}_\text{o}$ equals the state estimate. 
We want to construct another distribution $p_\text{a}(\vec{s}) \sim N(\vecsym{\mu}_\text{a}, \mat{H}_\text{a}^{-1})$ that well approximates the original distribution with a sparser factor graph topology.

We follow NFR~\cite{mazuran2016nonlinear} and minimize the Kullback-Leibler divergence (KLD) between the recovered distribution and the original distribution. More formally, we minimize
\begin{multline}
    \label{eq:kl_div}
    D_\text{KL} (p(\vec{s}) || p_\text{a} (\vec{s})) = \\
    \frac{1}{2} \left(\langle\mat{H}_\text{a}, \mat{\Sigma}_\text{o}\rangle - \log \det (\mat{H}_\text{a} \mat{\Sigma}_\text{o}) + || \mat{H}_\text{a}^{\frac{1}{2}} (\vecsym{\mu}_\text{a} - \vecsym{\mu}_\text{o})||^2 - d\right),
\end{multline}
where  $\mat{\Sigma}_\text{o} = \mat{H}_\text{o}^{-1}$ and $d$ is constant. 

For the $i$th non-linear factor that we want to recover, we need to define a residual function such that $\vec{r}_i(\vec{s}, \vec{z}_i) = \vecsym{\epsilon}$ with $\vecsym{\epsilon} \sim N(\vec{0}, \mat{H}_i^{-1})$. 
NFR estimates the pseudo measurements $\vec{z}_i$ and information matrices $\mat{H}_i$ for the factors.
Choosing $\vec{z}_i$ such that $\vec{r}_i( \vecsym{\mu}_\text{o}, \vec{z}_i ) = \vec{0}$ induces $\vecsym{\mu}_\text{a} = \vecsym{\mu}_\text{o}$ which makes the third term of \eqref{eq:kl_div} vanish. 
To estimate $\mat{H}_i$ we define 
\begin{align}
    \mat{J}_\text{r} =
    \begin{bmatrix}
        \vdots \\
        \mat{J}_i \\
        \vdots \\
    \end{bmatrix}
    \mat{H}_\text{r} =
    \begin{bmatrix}
        \ddots &  & 0 \\
         & \mat{H}_i &  \\
        0 &  & \ddots \\
    \end{bmatrix}\,,\quad
\end{align}
where $\mat{J}_\text{r}$ stacks the Jacobians of the defined residual functions with respect to the state, and $\mat{H}_\text{r}$ is a block diagonal matrix that consists of the $\mat{H}_i$ for the corresponding residual functions. 
This allows us to write $\mat{H}_\text{a} = \mat{J}^\top_\text{r} \mat{H}_\text{r} \mat{J}_\text{r}$, and consequently, we can recover the information matrices $\mat{H}_i$ by minimizing
\begin{multline}
    \label{eq:d_kl_error}
    D_\text{KL} (\mat{H}_\text{r}) =
    \langle \mat{J}^\top_\text{r} \mat{H}_\text{r} \mat{J}_\text{r}, \mat{\Sigma}_\text{o}\rangle - \log \det (\mat{J}^\top_\text{r} \mat{H}_\text{r} \mat{J}_\text{r}).
\end{multline}
For full-rank and invertible $\mat{J}_\text{r}$,~\cite{mazuran2016nonlinear, hsiung2018} showed that the following closed-form solution exists,
\begin{align}
    \label{eq:recovery_solution}
    \mat{H}_i = (\{\mat{J}_\text{r} \mat{\Sigma}_\text{o} \mat{J}_\text{r}^\top\}_i)^{-1},
\end{align}
where $\{\}_i$ denotes the corresponding diagonal block.

\begin{table*}[t]
\begin{center}
\resizebox{\textwidth}{!}{%
  \begin{tabular}{r |c c c c c c c c c c}
    \toprule
Sequence & MH\_01 & MH\_02 & MH\_03 & MH\_04 & MH\_05 & V1\_01 & V1\_02 & V1\_03 & V2\_01 & V2\_02 \\
\midrule

VI DSO \cite{stumberg18direct}, mono  & \textbf{0.06} & \textbf{0.04} & 0.12 & \textbf{0.13} & 0.12 & 0.06 & 0.07 & \textbf{0.10} & \textbf{0.04} & 0.06 \\
OKVIS \cite{leutenegger2014keyframe} mono  & 0.34 & 0.36 & 0.30 & 0.48 & 0.47 & 0.12 & 0.16 & 0.24 & 0.12 & 0.22 \\
OKVIS \cite{leutenegger2014keyframe} stereo  & 0.23 & 0.15 & 0.23 & 0.32 & 0.36 & \textbf{0.04} & 0.08 & 0.13 & 0.10 & 0.17 \\
VINS FUSION \cite{qin2019} mono  & 0.18 & 0.09 & 0.17 & 0.21 & 0.25 & 0.06 & 0.09 & 0.18 & 0.06 & 0.11 \\
VINS FUSION \cite{qin2019} stereo  & 0.24 & 0.18 & 0.23 & 0.39 & 0.19 & 0.10 & 0.10 & 0.11 & 0.12 & 0.10 \\
IS VIO \cite{hsiung2018} stereo  & \textbf{0.06} & 0.06 & 0.10 & 0.24 & 0.19 & 0.06 & 0.10 & 0.26 & 0.08 & 0.21 \\
\bf Proposed VIO, stereo  & 0.07 & 0.06 & \textbf{0.07} & \textbf{0.13} & \textbf{0.11} & \textbf{0.04} & \textbf{0.05} & \textbf{0.10} & \textbf{0.04} & \textbf{0.05} \\
\midrule
VI SLAM \cite{Kasyanov2017_VISLAM} mono,  KF  & 0.25 & 0.18 & 0.21 & 0.30 & 0.35 & 0.11 & 0.13 & 0.20 & 0.12 & 0.20 \\
VI SLAM \cite{Kasyanov2017_VISLAM} stereo, KF  & 0.11 & 0.09 & 0.19 & 0.27 & 0.23 & 0.04 & 0.05 & 0.11 & 0.10 & 0.18 \\
VI ORB-SLAM \cite{mur-artal-inertial}, mono, KF  & \textbf{0.07} & 0.08 & 0.09 & 0.22 & \textbf{0.08} & \textbf{0.03} & 0.03 & X & \textbf{0.03} & 0.04 \\
Pure BA, stereo, KF &  0.09 &   0.08  &   \textbf{0.05}  &   0.27  &  0.16  &   0.04  &  0.03  & X  &  0.04   &  0.04 \\
BA + Identity Factors, stereo, KF & 0.08 & 0.07 & X & 0.34 & 0.15 & 0.04 & 0.03 & 0.56 & 0.05  &  0.04 \\
\bf Proposed VI Mapping, stereo, KF  & 0.08 & \textbf{0.06} & \textbf{0.05} & \textbf{0.10} & \textbf{0.08} & 0.04 & \textbf{0.02} & \textbf{0.03} & \textbf{0.03} & \textbf{0.02} \\
\bottomrule
  \end{tabular}
}
\end{center}

\caption{RMS ATE of the estimated trajectory in meters on the EuRoC dataset for several different methods. In the upper part we summarize the results for the VIO methods that run optimization in a local window and estimate the pose of every camera frame. In the lower part we evaluate mapping methods that operate on all keyframes and perform global map optimization. In both evaluations the proposed system shows the lowest error on the majority of the sequences and outperforms the competitors.
Note: The V2\_03 sequence is excluded from the comparison because it has more than 400 missing frames for one of the cameras.
}
\vspace{-5mm}
\label{tab:euroc}

\end{table*}

\subsection{Non-Linear Factors for Distribution Approximation}

When we need to marginalize out a keyframe as shown in Fig. \ref{fig:factor_graph} (c), we save the current linearization and marginalize out everything except the keyframe poses. 
This gives us a factor that densely connects all keyframe poses in the optimization window.
We use it to recover non-linear factors between the marginalized keyframe and all other keyframes as shown in Fig. \ref{fig:recovered_graph}.
We define the following residual functions:
\begin{align}
\vec{r}_\text{rel}(\vec{s}, \vec{z}_\text{rel}) &= \logmap( \vec{z}_\text{rel}  \mat{T}_{j}^{-1} \mat{T}_{i}),\\
\vec{r}_\text{rp}(\vec{s}, \vec{z}_\text{rp}) &= \lfloor \vec{z}_\text{rp}  \rot_{i}^{-1} (0,0,-1)^\top \rfloor_{xy},\\
\vec{r}_\text{pos}(\vec{s}, \vec{z}_\text{pos}) &= \vec{z}_\text{pos} - \tran_{i},\\
\vec{r}_\text{yaw}(\vec{s}, \vec{z}_\text{yaw}) &= \lfloor \rot_{i} \vec{z}_\text{yaw} \rfloor_y,
\end{align}
where with $\lfloor \rfloor_{xy}$ we denote $x$ and $y$ components of the vector and with $\vec{z}$ we denote the recovered measurements from the estimated state at the time of linearization. In our case $\vec{z}_\text{rel}  = \mat{T}_{i}^{-1} \mat{T}_{j} \in \SE(3)$, $\vec{z}_\text{rp}= \rot_{i}  \in \SO(3) $, $\vec{z}_\text{pos}  = \tran_{i} \in \Real^3$ and $\vec{z}_\text{yaw} = \rot_{i}^{-1} \begin{pmatrix} 1 & 0 & 0 \end{pmatrix}^\top  \in \Real^3$.

We recover pairwise relative-pose factors between the keyframe that we will remove and all other current VIO keyframes. For that keyframe we also recover roll-pitch, absolute position and yaw factors (Fig. \ref{fig:recovered_graph}). This gives us a full-rank invertible Jacobian $\mat{J}_\text{r}$ which means that we can use Eq.~\eqref{eq:recovery_solution} for recovering information matrices for the factors. 

Since yaw and absolute position are 4 unobservable states of the VIO, the only information we have there comes from the initial prior on the start pose. As we do not need this information for the global map we drop yaw and absolute position factors, and only take relative pose and roll-pitch factors for the map optimization.
With these factors, the energy terms $E^\text{G}_{\text{nfr}}$ become
\begin{align}\label{eq:Enfr}
    E^\text{G}_{\text{nfr}}(\vec{s}) &= \sum_{\mathclap{\substack{(i,j)\in\mathcal{R}}}} \vec{r}_{ij}^\top \mat{H}_{ij} \vec{r}_{ij}
    +\sum_{\mathclap{\substack{i\in\mathcal{P}}}} \vec{r}_{i}^\top \mat{H}_{i} \vec{r}_{i},
\end{align}
where $\mathcal{R}$ is a set of all relative pose factors and $\mathcal{P}$ is the set of all roll-pitch factors.

\section{Evaluation}

To evaluate the presented approach we conduct evaluation on the EuRoC dataset \cite{Burri16} and compare it to other state-of-the-art systems. We present the evaluation for both our VIO subsystem and our full visual-inertial mapping approach. Our VIO runs the optimization in a local window of frames and provides a pose for every tracked frame, while the mapping system performs global map optimization for keyframes that were selected by the VIO. 
To measure the accuracy of the evaluated systems, we use the root mean square (RMS) of the absolute trajectory error (ATE) after aligning the estimates with ground truth.

\paragraph{System parameters} At the KLT tracking stage the image is divided into a regular grid with the cell size of 50 pixels. For each cell that has no point tracked from the previous frame, one feature point with the best FAST response is extracted (if it exceeds the threshold). With the resolution of the EuRoC dataset it results in 80-120 features tracked by the system at every point in time. At the VIO level we use a window of 7 old keyframes (poses) and 3 latest temporal states (poses, velocities and biases). The newest temporal state is selected as a keyframe if less than 70\% of the KLT features are connected to the currently tracked points in the local map.

\paragraph{Accuracy} The results of the evaluation are summarized in Table~\ref{tab:euroc}. 
When considering visual-inertial odometry methods our system shows the best performance on eight out of ten sequences while the closest competitor (VI DSO \cite{stumberg18direct}) shows the best results on five.

To evaluate the mapping part we compare it to the visual-inertial version of ORB-SLAM  \cite{mur-artal-inertial}, where the vision subsystem is very similar to the one proposed in our mapping layer (ORB keypoints). The main difference lies in the inertial part where ORB-SLAM uses preintegrated measurements between keyframes, while we use recovered non-linear factors that summarize IMU and visual tracking on the VIO layer.

The proposed system clearly outperform ORB-SLAM on the ``machine hall'' sequences where the large scale of the environment results in large time intervals between keyframes. 
On the ``Vicon room'' sequences the difference is smaller, since the rapid motion of the MAV that carries the camera in a small room results in many keyframes with small time intervals between them. 

Qualitative results of reconstructed maps are shown in Fig.~\ref{fig:teaser}. 
With the proposed system we are able to reconstruct globally consistent gravity-aligned maps and recover keyframe poses even for segments where no matches between detected ORB features can be estimated.

\paragraph{Factor Weighting} To evaluate the importance of the extracted factors and their proper weighting in the final mapping results we consider two alternative implementations. In the first one we do not use any factors and rely purely on the BA with ORB features. In the second one we extract the factors, but use identity weights (i.e. $\mat{H}_{ij}=\mat{H}_i=\mat{I}$ in Eq.~\eqref{eq:Enfr}) for all of them, which is a typical approach for pose graph optimization~\cite{mur-artal-inertial,qin2019}. The evaluation results presented in Table~\ref{tab:euroc} show that the system with the factor weights recovered according to Sec.~\ref{sec:vi_mapping} results in better accuracy and robustness when compared to those alternatives.

\paragraph{Timing} The main source of timing improvement for the mapping stage is the fact that for a global optimization requires a 2.5 smaller state (no velocity or biases) compared to the naive IMU integration. In absolute numbers we test our system on an Intel E5-1620 CPU (4 cores, 8 virtual cores). Our implementation is highly parallel and utilizes all available CPU resources. For the VIO the average time per frame on the EuRoC sequences is 7.83 ms (largest: 9.4 ms on MH\_02; smallest: 5.5 ms in V1\_03). On average 11.5\% of the frames are selected as keyframes and proceed to the mapping stage. 

The timing of the mapping stage is provided in Table \ref{tab:mapping_results}. In particular, for the MH\_05 sequence (see Fig. 1, 2273 stereo frames, 114 seconds) the processing takes 19.2 seconds for VIO and 9.7 seconds for mapping for the entire sequence (around 4x faster than real-time playback).

\begin{table}[t]
\resizebox{\linewidth}{!}{%
  \begin{tabular}{c c c c c}
    \toprule
    Total & \makecell{Factor \\ Extraction} & \makecell{Keypoint \\ detection} & \makecell{Matching and \\ Triangulation} & \makecell{Optimization \\ (10 iterations)} \\
    \midrule
    52.8 & 3.6 & 6.4 & 23.1 & 19.7 \\
\bottomrule
\end{tabular}
}
  \caption{Mean processing time in \textbf{milliseconds} of the mapping subsystem on EuRoC sequences normalized (divided) by the number of keyframes in the map.}
    \vspace{-5mm}
\label{tab:mapping_results}

\end{table}

\section{Conclusions}

In this paper we present a novel approach for visual-inertial mapping that combines the strengths of highly accurate visual-inertial odometry with globally consistent keyframe-based bundle adjustment. 
We achieve this in a hierarchical framework that successively recovers non-linear factors from the VIO estimate that summarize the accumulated inertial and visual information between keyframes.
VIO is formulated as fixed-lag smoothing which optimizes a set of active recent frames in a sliding window and keeps past information in marginalization priors. 
The accumulated VIO information between keyframes is extracted and retained for the visual-inertial mapping when a keyframe falls outside the window and is marginalized. 

Compared to alternative approaches that use preintegrated IMU measurements between keyframes our system shows better trajectory estimates on a public benchmark. 
This formulation has the potential to reduce the computational cost of optimization by reducing the dimensionality of the state space and enable large-scale visual-inertial mapping.
Integrating information from other sensor modalities or extending the system for multi-camera settings are interesting directions for future research.

{\small
\bibliographystyle{IEEEtran}
\bibliography{IEEEabrv,egbib}
}

\balance

\end{document}